\documentclass[11pt]{article}
\usepackage[margin=2.3cm]{geometry}
\usepackage{amsmath,amssymb}
\usepackage{graphicx}
\usepackage{booktabs}
\usepackage{array}
\usepackage{xcolor}
\usepackage{microtype}
\usepackage{caption}
\usepackage[numbers,sort&compress]{natbib}
\usepackage[colorlinks=true,linkcolor=blue!50!black,citecolor=blue!50!black,urlcolor=blue!50!black]{hyperref}
\newcommand{\kge}{KGE$^{\prime}$}
\newcommand{\recipe}{AIFL-recipe}
\newcommand{\rfc}{\mbox{KNOU-Flood}}
\graphicspath{{figs/}}
\title{\rfc: Advancing global streamflow forecasting with lead-aware probabilistic machine learning}
\author{Minjong Cheon\\[2pt]\small Department of Computer Science, Korea National Open University (KNOU)}
\date{}

\begin{document}
\maketitle

\begin{abstract}
Global streamflow forecasting supports flood preparedness, yet data-driven models trained on reanalysis lose skill when driven by operational forecasts, increasingly so with lead time. AIFL addresses this shift with a two-stage transfer-learning recipe and a deterministic LSTM, but it feeds one forecast lead time into the whole input window and gives no uncertainty. We ask how far this open recipe can be extended without changing its recurrent core. \rfc{} keeps a single-layer LSTM and the two-stage training and changes the information around it: a longer input window (424 days), lead-aware forecast inputs, GraphCast as a second forecast source next to ECMWF HRES, and a latent-noise ensemble trained with the fair CRPS. We compare against a re-implementation of the AIFL recipe (its code is not public) under identical data, stations, splits and metrics. Our contributions are threefold. First, over three seeds and 2{,}099 test basins (2021--2023), \rfc{} reaches a median \kge{} of 0.65 at LT1 and 0.48 at LT10 against 0.62 and 0.34, with significant gains at all ten lead times under two scoring protocols, also against a 1024-unit baseline and on held-out basins (one seed each). Second, identical weights give an LT10 \kge{} of 0.18 or 0.34 depending on how forecasts are fed, so much of the lead-time decay reported for AIFL is an input-protocol effect; the probabilistic head contributes most to the gain, followed by the raw GraphCast channels, the longer window and mixed fine-tuning. Third, the ensembles reach lower CRPS than probabilistic versions of the baseline, are nearly calibrated with inference-time dropout, and detect 1.5- and 2-year floods better than all baselines; the benefit for rarer floods depends on the seed, and an exploratory tail-weighted variant improves 10-year and larger events. \rfc{} is a controlled extension of an open recipe, not a new architecture.
\end{abstract}

\section{Introduction}
Floods are among the most frequent and damaging natural hazards, and their toll is growing: the number of major flood events recorded worldwide more than doubled between 1980--1999 and 2000--2019 \citep{undrr2020}. Timely forecasts of river discharge allow authorities and communities to prepare, but only if they are accurate over the lead times at which action is possible and honest about their uncertainty. Providing such forecasts everywhere, including in the large parts of the world where observation networks and hydrological expertise are thin, remains a central challenge for hydrology.

Global flood forecasting has traditionally relied on physically based hydrological models driven by numerical weather prediction (NWP), whose skill depends on laborious calibration and on the quality of the forcing \citep{harrigan2020}. Machine learning has begun to change this. Long short-term memory (LSTM) networks trained jointly on many basins learn hydrological behavior that transfers across catchments, outperform calibrated process-based models and extend to basins without observations \citep{kratzert2018,kratzert2019}. Google's flood forecasting system carried this approach to ungauged basins worldwide \citep{nearing2024,nevo2022}, and ECMWF's AIFL recently showed that a deliberately simple LSTM can be trained within the open Caravan ecosystem and driven by operational forecasts \citep{aifl2026}.

A structural problem arises at the moment such models are put to use. They learn rainfall--runoff relations from reanalysis, a best estimate of past weather, but in operation they are driven by forecasts whose errors grow with lead time and whose statistics differ from those of the reanalysis. Ignoring this reanalysis-to-forecast domain shift forfeits a substantial part of the skill \citep{biascast2025}. AIFL narrows the gap with a two-stage strategy: pre-training on 40 years of ERA5-Land and fine-tuning on the one-day-ahead forecast. Its authors identify two limits of this design. Every input step uses the one-day-ahead forecast, so the dependence of skill on lead time can be probed only by substituting a longer-lead forecast throughout the input window, which is not how forecasts are available in operation. And the model is deterministic: it offers no measure of uncertainty and, trained with a squared-error loss, underestimates peak discharge.

Here we ask how far this recipe can be extended without touching its recurrent core. \rfc{} retains the two-stage strategy and a single-layer LSTM and changes the information around it: a longer input window; forecast inputs that depend on lead time; a second forecast source, GraphCast \citep{lam2023}, alongside ECMWF HRES; and a probabilistic output, a latent-noise ensemble trained with the fair continuous ranked probability score (CRPS) \citep{ferro2014}. We keep the architecture deliberately simple so that every change can be attributed. Because the code of AIFL is not public, we compare against a re-implementation of its recipe, trained on identical stations, splits and metrics, so that differences reflect the design rather than the data. \rfc{} is not a new network architecture, and we make no claim that it is: like the AIFL recipe and the Google system, it rests on an LSTM, and what we contribute is a controlled extension and an analysis of what matters.

The paper makes three contributions. First, a controlled comparison with the AIFL recipe. Across three independently trained seeds and about 2{,}100 stations, \rfc{} reduces the error at every lead time; under operational scoring the gain in \kge{} grows from 0.04 on the first day to 0.14 on the tenth, and it persists against a larger baseline and, in a single-seed test, on stations held out in space. Second, an analysis of the scoring protocol and of the model components. A large share of the steep decay of skill with lead time reported for the AIFL protocol reflects how forecasts are fed to the model rather than the quality of the forecasts themselves, and a decomposition of the gain attributes most of it to the probabilistic head, followed by the raw GraphCast channels, the longer window and the mixed fine-tuning. Third, a probabilistic and event-based evaluation. The ensembles outperform probabilistic versions of the baseline and become nearly calibrated when dropout is kept active at inference, without retraining, and they identify floods of 1.5- to 2-year return period more reliably than every baseline. The benefit for rarer floods, however, varies with the training seed; a tail-weighted training objective raises it, but we introduced that objective after observing this variability in the test years and treat the result as exploratory.

We first review related work (Section~\ref{sec:related}), then describe the data and the experimental design (Section~\ref{sec:data}), the model and its training (Section~\ref{sec:method}) and the results (Section~\ref{sec:results}), and conclude with future directions (Section~\ref{sec:conclusion}).

\section{Related Work}\label{sec:related}
\paragraph{Machine learning for rainfall--runoff modeling.} LSTM networks \citep{kratzert2018} have become the standard data-driven rainfall--runoff model, and training one model on many basins with static attributes as additional input improves prediction in basins without observations \citep{kratzert2019}. Extensions add physical structure, for example mass conservation in the recurrent cell \citep{hoedt2021}, or represent the routing of flow along the river network. RiverMamba, a state-space model pre-trained on reanalysis, forecasts discharge on a global 0.05$^\circ$ grid up to seven days ahead and exploits the channel network \citep{shams2025}. These approaches trade simplicity for spatial realism. We remain with the lumped, station-by-station formulation of AIFL in order to isolate the effect of the training strategy and of the forecast inputs.

\paragraph{Global flood forecasting systems.} GloFAS couples NWP with a process-based hydrological model \citep{harrigan2020}. Google's system uses an LSTM with a hindcast encoder and a forecast decoder that receives the weather forecast at each lead time, and it predicts a distribution (an asymmetric Laplace) at every step \citep{nearing2024,nevo2022}. AIFL \citep{aifl2026} trades this specialized architecture for a deterministic single-layer LSTM trained end to end within Caravan \citep{kratzert2023caravan}. The Caravan-MultiMet extension \citep{multimet2024}, which adds archives of HRES and GraphCast forecasts to more than 22{,}000 basins, makes forecast-driven training and evaluation possible on a common basis.

\paragraph{Reanalysis-to-forecast domain shift.} Models trained on reanalysis lose skill when driven by forecasts. BiasCast reports that the median Nash--Sutcliffe efficiency in 451 central European basins falls from 0.58 to 0.33 at one-day lead when a reanalysis-trained model is driven by forecasts, and that a single LSTM which processes hindcast and forecast sequentially recovers most of the loss \citep{biascast2025}. AIFL instead fine-tunes a reanalysis-pre-trained network on forecasts \citep{aifl2026}. \rfc{} combines the two ideas, a sequential LSTM with lead-aware forecast inputs and the two-stage recipe, and extends them to ten lead times and two forecast sources.

\paragraph{Probabilistic streamflow forecasting.} Distributional output layers, such as mixture density networks, give calibrated predictive distributions for rainfall--runoff prediction \citep{klotz2022}, and the Google system predicts the parameters of an asymmetric Laplace distribution \citep{nearing2024}. Such heads describe the marginal distribution at each time step. Dropout \citep{gal2016} and generative models, for instance diffusion models with a state-space backbone \citep{hydrodiffusion2025}, produce samples or trajectories. In weather forecasting, ensembles trained directly on the CRPS now match or outperform physics-based ensembles at medium range \citep{lang2024}. \rfc{} adopts this idea in a lightweight form: latent noise injected late in the input window yields trajectory-consistent members at almost no extra cost, trained with the fair CRPS \citep{ferro2014}. We show that dropout alone gives strongly under-dispersive ensembles for streamflow, and that combining it with latent noise nearly calibrates the ensemble without retraining.

\section{Data Curation \& Experimental Design}\label{sec:data}
\begin{figure}[t]
\centering\includegraphics[width=0.96\textwidth]{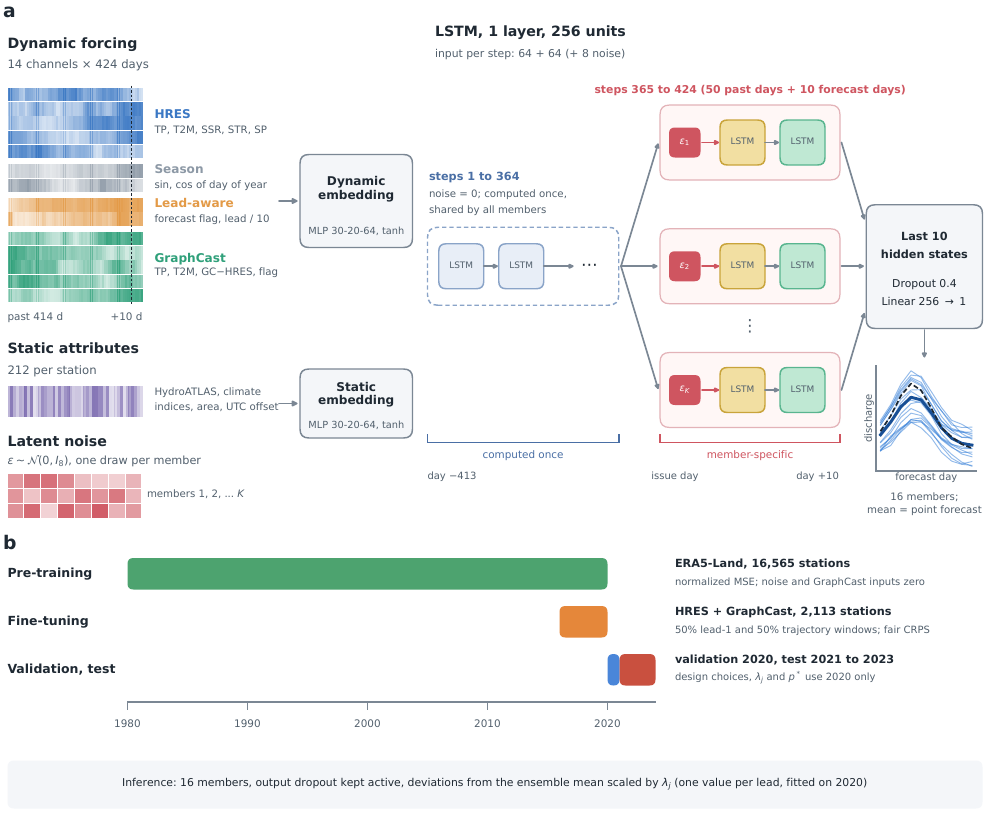}
\caption{Schematic of the \rfc{} framework. \textbf{a}, Separate MLP embeddings for dynamic and static inputs feed a shared LSTM core that processes a 424-day window (414 past days and 10 forecast days). Latent noise enters in the last 60 steps only; the first 364 steps are computed once and shared by all ensemble members. The hidden states of the last 10 steps give the discharge of the 10 forecast days for each member. \textbf{b}, The training strategy transitions from ERA5-Land reanalysis pre-training to HRES and GraphCast forecast fine-tuning; 2020 is used for all design decisions and 2021--2023 for testing.}\label{fig:framework}
\end{figure}

This section describes the data sources and the processing steps used to develop \rfc. The focus is on a globally consistent training set that supports a two-stage reanalysis-to-forecast learning strategy. The \rfc{} framework (Figure~\ref{fig:framework}) uses separate feedforward embedding networks for static landscape attributes and dynamic meteorological forcing, which are integrated by an LSTM core. Latent noise enters in the last part of the input window so that the model produces an ensemble of discharge trajectories.

\subsection{Datasets and Target Variable}
The model is trained to predict daily streamflow in the form of specific discharge (mm\,d$^{-1}$), using station-based observations from the Caravan dataset \citep{kratzert2023caravan} (core release v1.6) and its GRDC-Caravan extension, restricted to basins with ERA5-Land forcing. This yields 18{,}133 stations. Meteorological forcing data are provided by the Caravan-MultiMet extension \citep{multimet2024}, which supplies catchment-averaged daily series (one value per basin, variable and day) from ERA5-Land reanalysis \citep{era5land} (native resolution approximately 0.1$^\circ$), ECMWF IFS HRES forecasts (approximately 0.1$^\circ$) and GraphCast forecasts (0.25$^\circ$). Forecasts are indexed by valid date and lead time (LT1 to LT10). The HRES archive starts in 2016, and the GraphCast archive spans 2016-01-02 to 2023-12-21.

Specific discharge simplifies the learning task by aligning the units of the target with meteorological inputs such as precipitation, while basin size enters as an explicit static attribute.

\subsection{Station Selection and Quality Control}
\begin{figure}[t]
\centering\includegraphics[width=\textwidth]{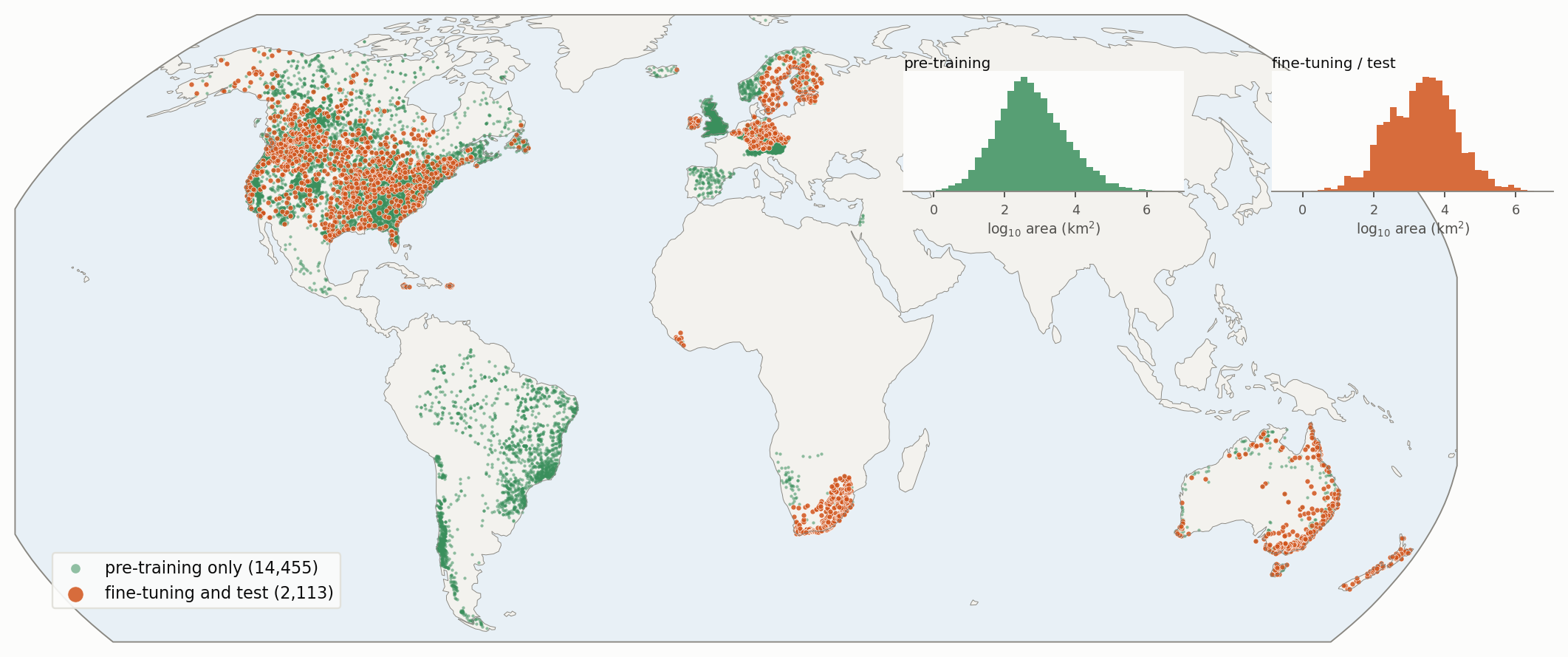}
\caption{Global spatial distribution of the streamflow stations across the experimental stages: pre-training only and fine-tuning/test. The inset diagrams provide the frequency distribution of catchment areas (log$_{10}$ scale) for the pre-training and the fine-tuning/test subsets.}\label{fig:stagemap}
\end{figure}

Pre-training uses all stations with at least three years of observations in 1980--2019 (16{,}565 stations; Figure~\ref{fig:stagemap}). Fine-tuning and testing use the 2{,}113 stations that have at least 365 observed days both in 2016--2019 and in the test window, following the AIFL design in which the same stations are fine-tuned and tested. We do not apply the polygon-overlap de-duplication or the flatline filters of AIFL. Because the baseline and \rfc{} are trained and evaluated on exactly the same stations, this choice does not affect the fairness of their comparison; it does mean that absolute scores are not directly comparable with those published for AIFL.

\subsection{Model Inputs and Consistency}
The model relies on two types of inputs: dynamic features, which vary over time, and static features, which are constant for each catchment (Table~\ref{tab:inputs}). The five core meteorological drivers (surface net solar radiation, surface net thermal radiation, surface pressure, 2\,m air temperature and total precipitation) are available in both ERA5-Land and HRES, which ensures that the input space is identical between pre-training and operational fine-tuning. Beyond these, \rfc{} receives (i) sine and cosine of the day of year, (ii) two lead-aware channels (a forecast flag and the lead time divided by ten), and (iii) five GraphCast channels: precipitation, temperature, their GraphCast-minus-HRES differences and an availability flag. During pre-training the GraphCast and lead-aware channels are set to zero. Static features consist of 212 catchment attributes (HydroATLAS \citep{linke2019} physiography, soil, geology, land cover and climate indices, Caravan climate indices, log catchment area and the UTC offset of the station). All inputs are z-score normalized with global training statistics.

\begin{table}[t]
\centering\small
\caption{Input features of \rfc. The five core variables are shared between ERA5-Land reanalysis and HRES forecasts in the Caravan-MultiMet dataset; the additional channels are described in Section~\ref{sec:method}.}\label{tab:inputs}
\begin{tabular}{>{\raggedright\arraybackslash}p{1.8cm}>{\raggedright\arraybackslash}p{6.3cm}>{\raggedright\arraybackslash}p{2.0cm}>{\raggedright\arraybackslash}p{3.6cm}}
\toprule
Variable & Description & Unit & Source\\
\midrule
SSR & Surface net solar radiation & W\,m$^{-2}$ & ERA5-Land, HRES\\
STR & Surface net thermal radiation & W\,m$^{-2}$ & ERA5-Land, HRES\\
SP & Surface pressure & kPa & ERA5-Land, HRES\\
T2M & Air temperature at 2 meters & $^\circ$C & ERA5-Land, HRES\\
TP & Total daily precipitation & mm\,d$^{-1}$ & ERA5-Land, HRES\\
DOY & Sine and cosine of day of year & -- & derived\\
LEAD & Forecast flag and lead time$/10$ & -- & derived\\
GC & GraphCast TP, T2M, GraphCast$-$HRES differences, availability & as above & GraphCast\\
Static & 212 catchment attributes (incl.\ log area, UTC offset) & various & Caravan, HydroATLAS\\
\bottomrule
\end{tabular}
\end{table}

\subsection{Data Availability for Temporal Evaluation}
\begin{figure}[t]
\centering\includegraphics[width=\textwidth]{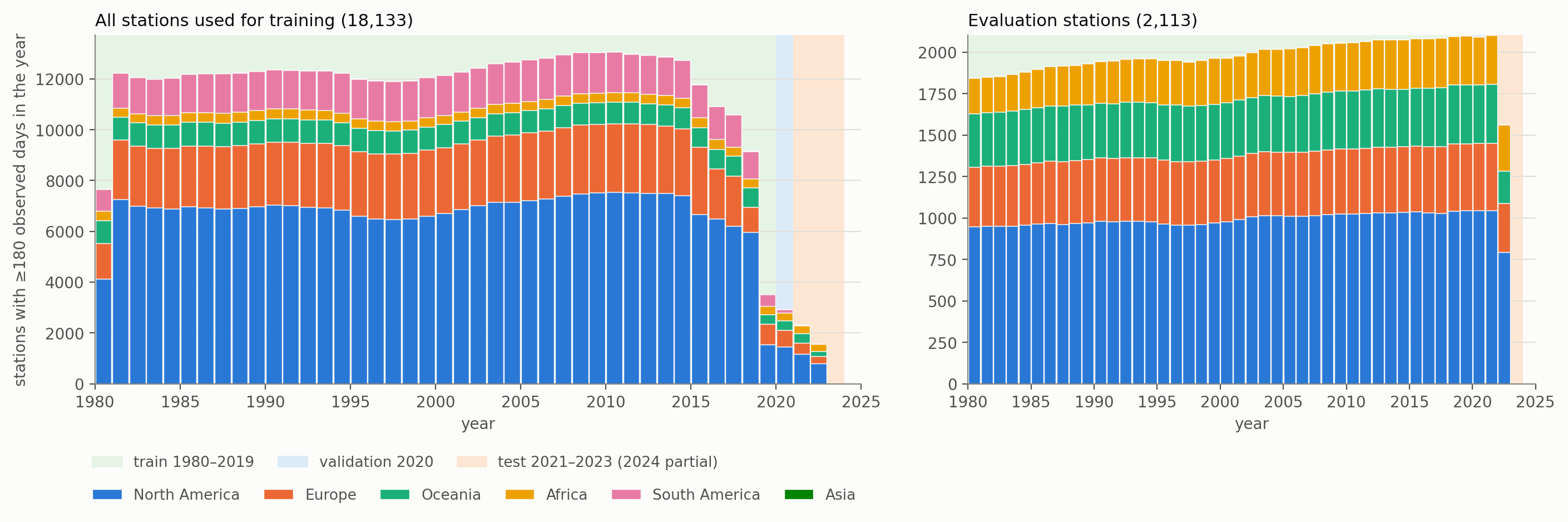}
\caption{Global station availability over time (1980--2023), by continent, for all stations (left) and for the evaluation stations (right). Shaded regions indicate the splits for pre-training and fine-tuning (green), validation (blue) and testing (orange). The drop after 2018 is why only about 2{,}100 stations can be scored in the test years.}\label{fig:avail}
\end{figure}

This study focuses on temporal generalization, evaluating the model's ability to forecast future events at established gauging stations. The decision to maintain a consistent set of stations, rather than a spatial split, is driven primarily by data availability: as shown in Figure~\ref{fig:avail}, there is a sharp decline in active station records after 2018, as many constituent datasets of Caravan lack updates beyond that date. The 2{,}113 stations available for the independent temporal test (2021 to the end of the GraphCast archive, with about 2{,}100 scored) are skewed toward larger catchments: their median area is 2{,}537\,km$^2$ against 488\,km$^2$ for the pre-training stations, and 25\% of them exceed 10{,}000\,km$^2$ against 11\%. We additionally report one experiment in which 20\% of the stations, selected as whole $2^\circ\times2^\circ$ blocks, are removed from both pre-training and fine-tuning and then used as test stations (435 stations), to assess the transfer to basins unseen in training; this experiment uses a single seed per model.

\section{Methodology}\label{sec:method}
This section describes the architecture and the training strategy of \rfc, the AIFL-recipe baseline and the scoring protocols. The code is written in PyTorch, with the data kept on the GPU, and is based on the AIFL recipe rather than on an existing framework.

\subsection{Model Architecture}
\rfc{} employs a single-layer LSTM network to capture temporal rainfall--runoff dynamics across a 424-day window comprising 414 past days and the 10 forecast days. The window was extended from the 180 days of the baseline in the validation experiments (Section~\ref{sec:components}), where it improved \kge{} and NSE at all ten lead times. Dynamic and static features are first transformed through separate three-layer feedforward embedding networks (layer sizes 30, 20, 64) with $\tanh$ activations. The embeddings are concatenated and fed into the LSTM, and the hidden states of the last 10 steps pass through a dropout layer (0.4) and a linear layer to give the 10-day discharge sequence. The architectural specifications are summarized in Table~\ref{tab:arch}.

\begin{table}[t]
\centering\small
\caption{Summary of the \rfc{} model architecture and hyperparameter choices.}\label{tab:arch}
\begin{tabular}{p{4.8cm}p{9.6cm}}
\toprule
Component & Configuration\\
\midrule
Recurrent layer & Single-layer LSTM, forget-gate bias initialized to 3.0\\
Hidden state size & 256\\
Dropout (output) & 0.4\\
Static embedding & 3-layer MLP (30, 20, 64) with $\tanh$\\
Dynamic embedding & 3-layer MLP (30, 20, 64) with $\tanh$\\
Temporal window & 424 days (414 past days and 10 forecast days)\\
Output sequence length & 10 days (hidden states of the last 10 steps)\\
Input features & 14 dynamic channels (Table~\ref{tab:inputs}) + 212 static attributes\\
Stochastic component & Latent noise $\varepsilon\sim\mathcal N(0,I_8)$ per ensemble member, concatenated to the LSTM input in the last 60 steps; steps 1--364 are computed once and shared; $K=8$ members in training, $K=16$ at inference\\
LSTM input per step & 136 (64 dynamic + 64 static + 8 noise)\\
Parameters & 0.41\,M\\
\bottomrule
\end{tabular}
\end{table}

The latent noise lets the model produce an ensemble at the cost of little more than a single forward pass, because the first 364 steps are computed once and shared by all members. At inference, the output dropout of the trained network may additionally be kept active (Monte Carlo dropout), which adds output-layer uncertainty that the latent noise alone does not cover. A single scalar $\lambda_j$ per lead time, fitted on the validation year by minimizing the CRPS, then scales the deviations of the members from their mean; the ensemble mean, which is the point forecast, is unchanged. No hyperparameter search was performed on the test data: all design decisions, including the architecture, were taken on the validation year 2020.

\subsection{Training Strategy}
\begin{figure}[t]
\centering\includegraphics[width=\textwidth]{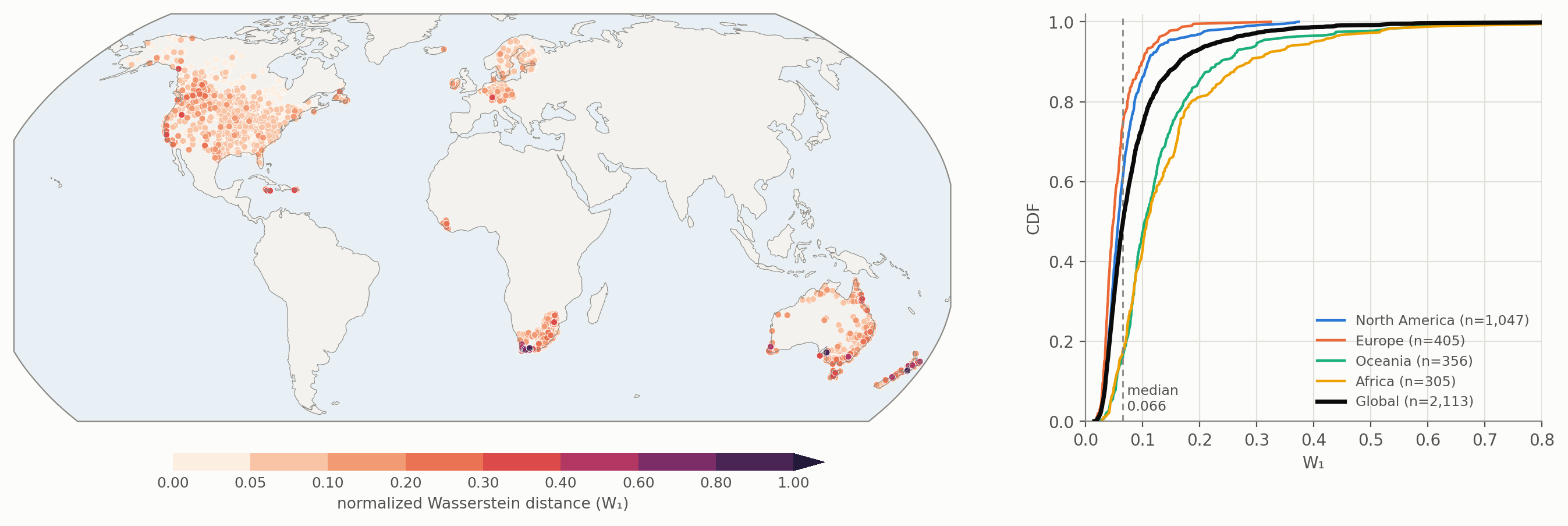}
\caption{Global spatial distribution of the normalized Wasserstein distance ($W_1$) between ERA5-Land reanalysis and LT1 HRES daily precipitation, computed for the 2{,}113 stations over 2016--2023. $W_1$ is the Earth Mover's Distance between the precipitation distributions, normalized by mean daily precipitation, yielding a dimensionless measure of relative distributional shift. The right panel shows the cumulative distribution of $W_1$ by continent.}\label{fig:wass}
\end{figure}

Systematic distributional shifts exist between the ERA5-Land reanalysis used for model development and the operational forcings encountered in real-time forecasting. As illustrated by the normalized Wasserstein distance in Figure~\ref{fig:wass}, the discrepancies between ERA5-Land and HRES precipitation are substantial across large parts of the globe, with a global median of 0.066 and a heavier tail in Africa and Oceania. Without an explicit alignment phase, a model trained solely on reanalysis would propagate these inconsistencies into its streamflow simulations. \rfc{} therefore adopts the two-stage transfer-learning strategy of AIFL (Table~\ref{tab:design}).

\paragraph{Pre-training.} The network is first optimized on ERA5-Land inputs over 1 January 1980 to 31 December 2019 across the 16{,}565 pre-training stations, using a normalized squared-error loss that weights each basin by its discharge variability,
\begin{equation}
\mathcal{L}_{\text{pre}}=\frac{1}{N}\sum_{i=1}^{N}\frac{(y_i-\hat y_i)^2}{\left(\sigma_{\text{basin}(i)}+\varepsilon\right)^{2}},\qquad \varepsilon=0.1,
\end{equation}
where $y_i$ and $\hat y_i$ denote observed and predicted specific discharge and $\sigma_{\text{basin}(i)}$ is the standard deviation of the observed discharge over the training period. Training uses the Adam optimizer with an initial learning rate of $5\times10^{-4}$, gradient clipping at 1.0 and a cosine schedule with a warm-up over the first 10\% of the updates, for 60 epochs of 3{,}000 updates with a batch size of 512. The latent-noise and GraphCast inputs are zero in this stage, so that a deterministic pre-trained checkpoint can be extended by zero-padding the corresponding input weights.

\paragraph{Fine-tuning.} Following pre-training, all weights are fine-tuned on HRES and GraphCast forecasts over 2016--2019 for 30 epochs of 1{,}500 updates at a reduced learning rate of $1.25\times10^{-4}$. Two kinds of windows are mixed in equal proportion: \emph{LT1 windows}, in which every input step uses the LT1 forecast as in AIFL, and \emph{trajectory windows}, in which the past days use LT1 forecasts and the forecast day $t+j$ uses the lead-$j$ forecast, i.e.\ the information that is actually available at the time of issue. The loss is the fair ensemble CRPS \citep{ferro2014}, a strictly proper score \citep{gneiting2007} corrected for the finite ensemble size, weighted as above,
\begin{equation}
\mathcal{L}_{\text{ft}}=\frac{1}{N}\sum_{i=1}^{N}\frac{1}{\sigma_{\text{basin}(i)}+\varepsilon}\left(\frac1K\sum_{k=1}^{K}\left|\hat y_{i,k}-y_i\right|-\frac{1}{2K(K-1)}\sum_{k\neq l}\left|\hat y_{i,k}-\hat y_{i,l}\right|\right),
\end{equation}
with $K=8$ members. Mixing both window types removes the lead-1 deficit that trajectory-only training exhibited on the validation year, and the same weights can then be scored under both protocols (Section~\ref{sec:protocols}). To ensure numerical continuity, the input scaler of the pre-training stage is reused.

We verified training stability by repeating the entire procedure, pre-training included, with three different seeds. The median LT1 \kge{} of the three seeds is 0.655--0.664 for \rfc{} and 0.615--0.634 for the baseline (AIFL protocol), which is small relative to the differences between the models.

\begin{table}[t]
\centering\small
\caption{\rfc{} experimental design and operational periods.}\label{tab:design}
\begin{tabular}{>{\raggedright\arraybackslash}p{2.3cm}>{\raggedright\arraybackslash}p{4.8cm}>{\raggedright\arraybackslash}p{2.4cm}>{\raggedright\arraybackslash}p{4.4cm}}
\toprule
Phase & Input data source & Period & Primary purpose\\
\midrule
Pre-training & ERA5-Land reanalysis & 1980--2019 & Learn universal hydrology\\
Fine-tuning & HRES + GraphCast (LT1 and trajectory windows) & 2016--2019 & Operational alignment\\
Validation & HRES + GraphCast & 2020 & Design decisions, calibration of $\lambda$ and $p^*$\\
Temporal test & HRES + GraphCast & 2021--2023-12 & Operational performance\\
\bottomrule
\end{tabular}
\end{table}

\subsection{AIFL-Recipe Baseline and Scoring Protocols}\label{sec:protocols}
The code, weights and station list of AIFL are not public. We therefore re-implemented its recipe from the paper (single-layer LSTM with the same embeddings and output dropout, a 180-day window, a normalized squared-error loss, pre-training on ERA5-Land, fine-tuning with LT1 forcing in all steps) and trained it on the same stations and splits as \rfc. This \emph{\recipe{} baseline} differs from the original in training budget (about 18\% of the pre-training updates), hidden size (256, with a 1024-unit control), the absence of de-duplication and quality filtering, no target noise and a different loss denominator (Table~\ref{tab:diff}). Because the baseline and \rfc{} share data, stations, splits and metrics, the comparison is controlled; the absolute values are not directly comparable with the published ones.

\begin{table}[t]
\centering\small
\caption{Published AIFL, the \recipe{} baseline and \rfc.}\label{tab:diff}
\begin{tabular}{>{\raggedright\arraybackslash}p{3.1cm}>{\raggedright\arraybackslash}p{3.7cm}>{\raggedright\arraybackslash}p{3.7cm}>{\raggedright\arraybackslash}p{3.9cm}}
\toprule
 & AIFL (paper) & \recipe{} baseline & \rfc\\
\midrule
Hidden size & 1024 & 256 (1024 control) & 256\\
Window & 180 d & 180 d & 424 d (414 + 10)\\
Forecast source & IFS control, LT1 in all steps & HRES, LT1 in all steps & HRES + GraphCast, lead-aware\\
Output & deterministic & deterministic & ensemble ($K=16$)\\
Loss (fine-tuning) & normalized MSE & normalized MSE & fair CRPS\\
Fine-tuning windows & LT1 everywhere & LT1 everywhere & 50\% LT1, 50\% trajectory\\
Pre-training updates & $\le$100\,epochs$\times$10\,k & 60$\times$3\,k & 60$\times$3\,k\\
Fine-tuning updates & $\le$30$\times$10\,k & 30$\times$1.5\,k & 30$\times$1.5\,k\\
Stations (pre / test) & 18{,}588 / 2{,}003 & 16{,}565 / 2{,}113 & 16{,}565 / 2{,}113\\
Parameters & $\approx$4.7\,M (our count) & 0.41\,M & 0.41\,M\\
\bottomrule
\end{tabular}
\end{table}

\paragraph{Scoring protocols.} The \emph{AIFL protocol} scores lead time $k$ by feeding the LT$k$ forecast into every input step with the same weights; this is our reading of the lead-sensitivity test of \citet{aifl2026}, whose description does not spell out the replacement, and our re-implementation reproduces the published pattern (NSE below zero beyond LT5). The \emph{operational (trajectory) protocol} scores a forecast issued on day $t$ using the LT1 forecasts of the past days and, for the forecast day $t+j$, the LT$j$ forecast.

\subsection{Evaluation Metrics}
Per station we compute the modified Kling--Gupta efficiency \kge{} \citep{kling2012}, the Nash--Sutcliffe efficiency (NSE) \citep{nash1970} and the RMSE in mm\,d$^{-1}$; the climatological RMSE (0.715\,mm\,d$^{-1}$, median) is reported alongside. Medians over stations are reported, and differences between systems are tested with a paired bootstrap over stations (2{,}000 resamples; 95\% interval of the difference of medians) applied to per-station metrics averaged over seeds. No post-hoc bias or damping correction is applied to the point forecasts.

Probabilistic forecasts are scored with the fair CRPS in physical units, the CRPS skill score relative to a day-of-year climatology (1980--2015, $\pm$15 days), rank histograms, the spread--error ratio, the coverage of the central interval and the fraction of observations outside the ensemble range. A deterministic forecast cannot be compared with an ensemble CRPS directly; we therefore convert the baseline into 16-member ensembles in two ways: (i) Monte Carlo dropout with the same $\lambda_j$ fitting as above, and (ii) residual dressing with the quantiles of the station- and lead-specific forecast residuals of 2020 in square-root flow space.

Flood events are defined relative to return-period thresholds (1.5 to 50 years) obtained by fitting a Gumbel distribution with L-moments \citep{hosking1990} to annual maxima. Observed events are identified when discharge exceeds the threshold derived from the observed annual maxima (stations with at least 10 years). Under a \emph{single-threshold} framework the forecast events use the same observed thresholds; under a \emph{dual-threshold} framework they use the thresholds derived from the annual maxima of the model's own ERA5-Land-driven simulation (1980--2024). Probability of detection (POD), false alarm ratio (FAR) and precision ($1-$FAR) are computed under an exact-day criterion, pooled over all stations and test days; the critical success index (CSI) is used to summarize POD and precision. A member-based event forecast issues an event when at least a fraction $p^*$ of the members exceeds the threshold, where $p^*$ is tuned per system, lead time and return-period group on the validation year only.

\section{Results and Evaluation}\label{sec:results}
This section evaluates the predictive skill of \rfc{} with a focus on temporal generalization, lead-time dependence, probabilistic skill and flood-event detection, always against the \recipe{} baseline. All results are reported for the temporal test setting, unless stated otherwise.

\subsection{Temporal Generalization and Global Performance}
\begin{figure}[t]
\centering\includegraphics[width=\textwidth]{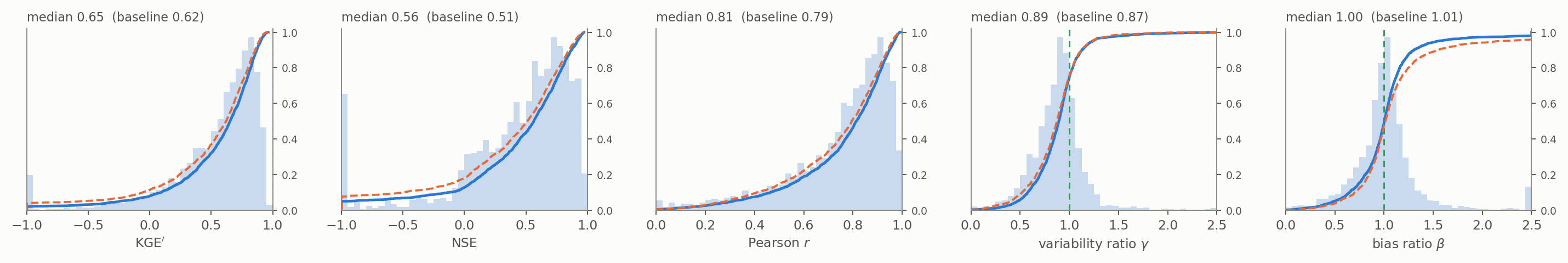}
\caption{Global performance metrics of \rfc{} across the test stations (LT1, AIFL protocol, mean of three seeds). Gray-blue histograms show the station frequency distribution and blue curves the empirical cumulative distribution function (CDF); orange dashed curves show the CDF of the \recipe{} baseline and green dashed lines the ideal value. Values are clipped to the axis range. From left to right: \kge{}, NSE, Pearson correlation $r$, variability ratio $\gamma$ and bias ratio $\beta$.}\label{fig:metrics}
\end{figure}

\rfc{} is evaluated over an independent temporal test period spanning 1 January 2021 to the end of the GraphCast archive (2023-12), at about 2{,}100 stations (2{,}099 with a finite score) selected solely by the availability of observations. Table~\ref{tab:main} compares the ensemble mean of \rfc{} with the baseline over three seeds trained from scratch. Under the AIFL protocol at LT1, \rfc{} achieves a median \kge{} of 0.65 and a median NSE of 0.56 (baseline: 0.62 and 0.51); for context, AIFL reports 0.66 and 0.53 at 2{,}003 other basins, which cannot be compared directly with our values owing to the different station list, period, filtering and training budget. The model maintains a high median Pearson correlation ($r=0.81$) and a median bias ratio $\beta$ of 1.00 (Figure~\ref{fig:metrics}). In both protocols \rfc{} is significantly better in \kge{}, NSE and RMSE at all ten lead times, and no lead time is worse. The advantage is modest at LT1 and grows with lead time: under the operational protocol the median RMSE at LT10 is 0.57 against 0.71\,mm\,d$^{-1}$ (climatology 0.715), and the fraction of stations that beat climatology is 89\% against 55\%.

\begin{table}[t]
\centering\small
\caption{Median over stations (test 2021--2023, three seeds; per-station metrics averaged over seeds). Lower is better for RMSE (mm\,d$^{-1}$); climatology 0.715. Differences (\rfc{} $-$ baseline, operational protocol; difference of medians with a 95\% paired-bootstrap interval) for LT1 / LT5 / LT10: \kge{} $+0.037$ [0.025, 0.045] / $+0.091$ [0.079, 0.101] / $+0.141$ [0.130, 0.154]; NSE $+0.052$ [0.038, 0.071] / $+0.165$ [0.143, 0.183] / $+0.281$ [0.267, 0.300]; RMSE $-0.032$ [$-$0.040, $-$0.015] / $-0.074$ [$-$0.087, $-$0.057] / $-0.141$ [$-$0.158, $-$0.116].}\label{tab:main}
\begin{tabular}{llcccccc}
\toprule
& & \multicolumn{3}{c}{\rfc} & \multicolumn{3}{c}{\recipe{} baseline}\\
Protocol & Metric & LT1 & LT5 & LT10 & LT1 & LT5 & LT10\\
\midrule
Operational & \kge{} & \textbf{0.655} & \textbf{0.550} & \textbf{0.481} & 0.618 & 0.459 & 0.340\\
 & NSE & \textbf{0.560} & \textbf{0.373} & \textbf{0.304} & 0.508 & 0.208 & 0.023\\
 & RMSE & \textbf{0.464} & \textbf{0.522} & \textbf{0.567} & 0.496 & 0.597 & 0.708\\
AIFL-style & \kge{} & \textbf{0.653} & \textbf{0.405} & \textbf{0.246} & 0.620 & 0.334 & 0.178\\
 & NSE & \textbf{0.564} & \textbf{0.087} & \textbf{$-$0.033} & 0.512 & $-$0.035 & $-$0.225\\
 & RMSE & \textbf{0.463} & \textbf{0.679} & \textbf{0.785} & 0.494 & 0.759 & 0.898\\
\bottomrule
\end{tabular}
\end{table}

\begin{figure}[t]
\centering\includegraphics[width=\textwidth]{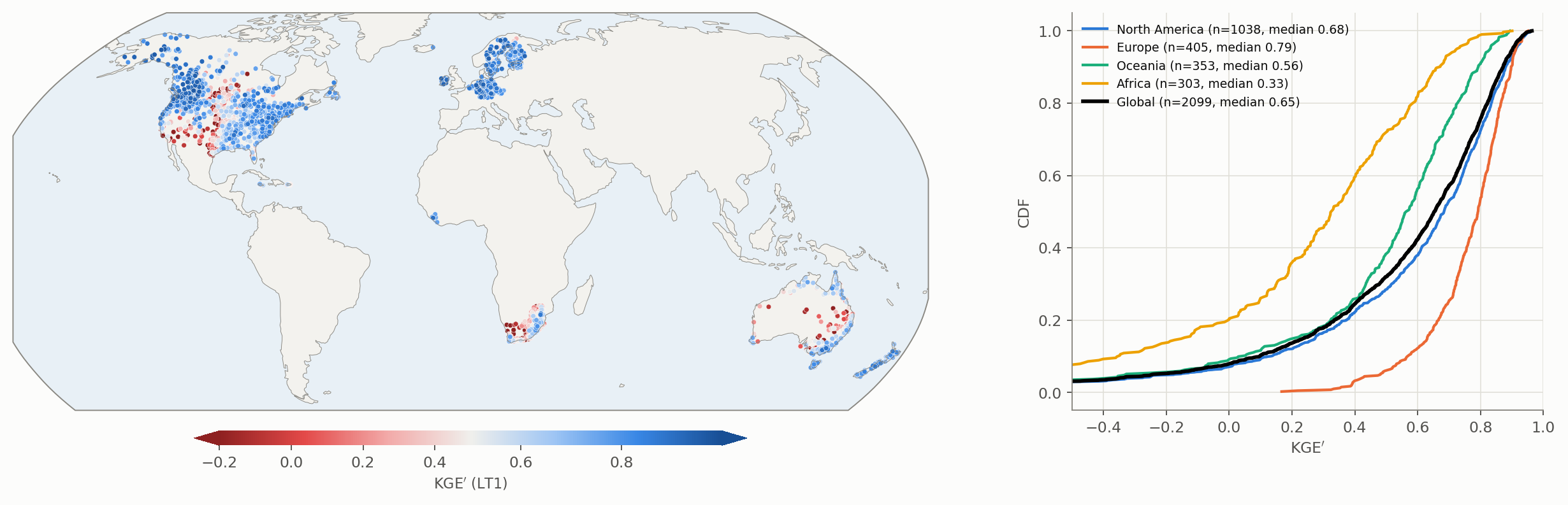}
\caption{Global and regional forecast performance (\kge{}) of \rfc{} for the 2021--2023 test period (LT1, AIFL protocol, mean of three seeds). The left panel shows the station-level \kge{}; the right panel shows the cumulative distribution of \kge{} by continent.}\label{fig:kgemap}
\end{figure}

The spatial distribution of \kge{} is shown in Figure~\ref{fig:kgemap}. Skill is consistently high across Europe and North America (median \kge{} 0.79 and 0.68), lower in Oceania (0.56) and lowest in Africa (0.33), where arid and intermittent-flow basins are concentrated. These catchments exhibit long periods of near-zero discharge punctuated by sparse sharp peaks, which strongly penalize composite metrics such as \kge{}.

\subsection{Lead-Time Dependence and Scoring Protocol}
\begin{figure}[t]
\centering\includegraphics[width=\textwidth]{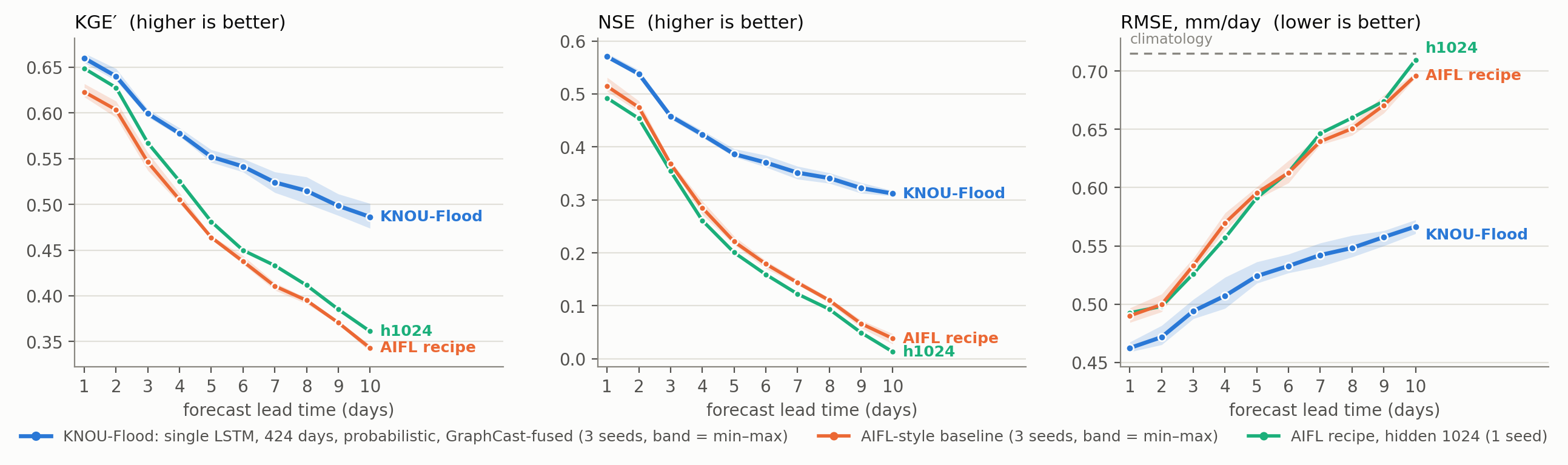}
\caption{\rfc{} performance as a function of lead time under the operational protocol, calculated across the test stations (median over 2{,}113 stations, 2021--2023). Median \kge{}, NSE and RMSE are shown for \rfc{} (three seeds; bands are min--max over seeds), the \recipe{} baseline (three seeds) and the baseline enlarged to 1024 units (one seed).}\label{fig:lead}
\end{figure}

Figure~\ref{fig:lead} shows the skill as a function of lead time under the operational protocol, together with a baseline enlarged to 1024 units. \rfc{} degrades much more slowly than the baselines.

\begin{figure}[t]
\centering\includegraphics[width=\textwidth]{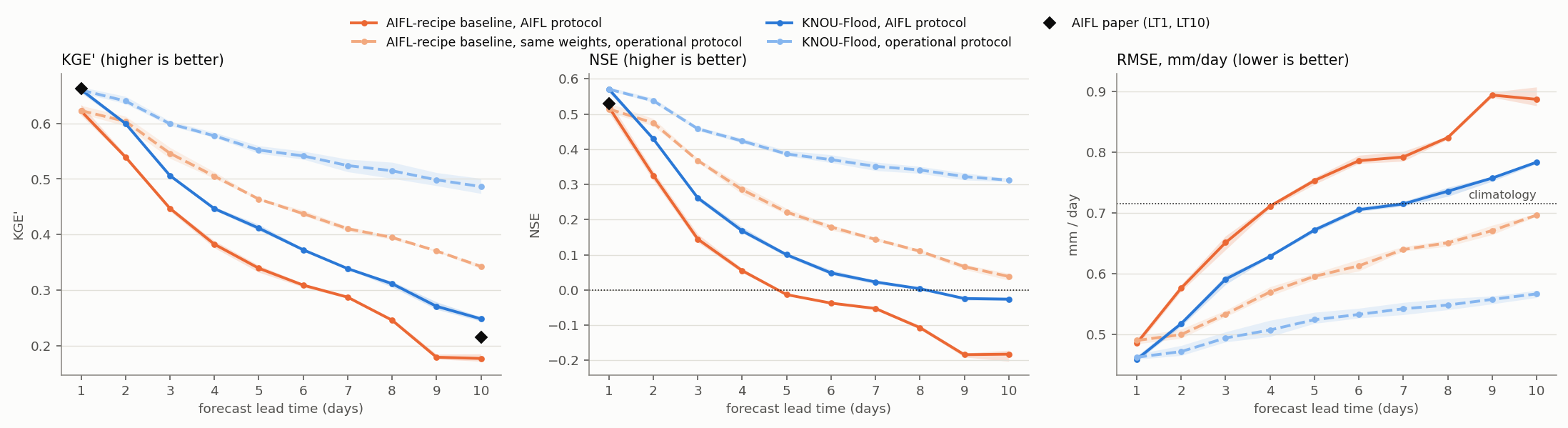}
\caption{Same weights, two scoring protocols (test, three seeds; bands are min--max over seeds). The AIFL protocol feeds the LT$k$ forecast into the whole input window; the operational protocol uses LT1 forecasts for the past days and each forecast day's own lead time. Diamonds mark the published AIFL values at LT1 and LT10. The AIFL-protocol curves follow our reading of AIFL's lead-sensitivity test, which the paper does not spell out.}\label{fig:protocol}
\end{figure}

To quantify the effect of the scoring protocol, we scored identical weights under both protocols (Figure~\ref{fig:protocol}). For the baseline, the LT10 \kge{} is 0.18 when the LT10 forecast fills the whole input window, which is close to the 0.215 published for AIFL, and 0.34 when only the forecast days use it; NSE changes from $-0.23$ to $+0.02$ and RMSE from 0.90 to 0.71\,mm\,d$^{-1}$. The monotonic degradation reported for the AIFL protocol therefore mixes forecast quality with an input design that replaces 180 days of meteorology by degraded forecasts, and the operational skill at long lead times is substantially higher than the protocol suggests. Because the original description of this test is not explicit, we flag this reading as ours.

\subsection{Controls: Model Capacity and Basins Unseen in Training}
A critic may argue that the baseline is simply too small. Enlarging it to 1024 units (one seed) raises the LT1 \kge{} from 0.618 to 0.648 but leaves NSE and RMSE essentially unchanged (NSE 0.516 to 0.500). \rfc{} (256 units, seed 0) remains significantly better than the larger baseline in \kge{}, NSE and RMSE at all ten lead times (operational protocol, LT1/LT5/LT10 \kge{}: 0.665/0.560/0.501 against 0.649/0.481/0.361). Conversely, enlarging a variant of our model to 1024 units did not improve it (NSE worse at all ten lead times), so we retained 256 units. Importantly, the 1024-unit baseline was trained with the same reduced update budget, and a training budget comparable to that of AIFL was not tested.

In the spatial hold-out experiment (435 stations; one seed per model), the advantage persists: LT1/LT5/LT10 \kge{} is 0.466/0.377/0.325 against 0.433/0.316/0.198, NSE is better at all ten lead times, \kge{} at seven and RMSE at eight (operational protocol), and the fraction of stations beating climatology at LT10 is 64\% against 36\%. Absolute skill is clearly lower than for stations seen in training.

\subsection{Component Analysis}\label{sec:components}
Table~\ref{tab:comp} summarizes the decomposition of the gain on the validation year (mostly one seed; magnitudes are indicative). The probabilistic CRPS head is the largest single contributor, followed by the GraphCast channels, the longer window and the mixed fine-tuning. A two-timescale design, with a weekly slow LSTM over 52 weeks coupled to a daily fast LSTM, was explored but is not better than a single LSTM with the same window: its apparent gain was the longer window.

\begin{table}[t]
\centering\small
\caption{Component analysis on the validation year 2020 unless stated otherwise (\kge{}, NSE and RMSE tested per lead time with a paired bootstrap).}\label{tab:comp}
\begin{tabular}{>{\raggedright\arraybackslash}p{5.2cm}>{\raggedright\arraybackslash}p{9.3cm}}
\toprule
Component & Outcome\\
\midrule
Probabilistic head (fair CRPS) versus deterministic & \kge{} better at 10/10 lead times, $+0.018$ to $+0.041$; the largest effect\\
GraphCast channels & \kge{} better at 10/10 lead times (deterministic single-LSTM comparison)\\
Window 180 $\to$ 424 days & \kge{} and NSE better at 10/10 lead times, about $+0.02$\\
Mixed LT1/trajectory fine-tuning & Removes the LT1 deficit of trajectory-only training (LT1 \kge{} 0.614 $\to$ 0.650)\\
Weekly slow clock (52 weeks) versus single LSTM, same window & No benefit: \kge{} 0/10 lead times; NSE better for the single LSTM at 5/10\\
Hidden size 1024 (two-timescale variant) & Worse NSE at 10/10 lead times\\
Day-of-year channel, lead-weighted loss, nowcast precipitation, discharge anchor & No or marginal effect (nowcast: $+$0.005 to 0.009 at LT1--LT5)\\
GraphCast$-$HRES difference channels removed (test, three seeds) & Negligible effect: \kge{} and NSE differences of at most $+0.009$, significant at 1 of 10 (operational) and 4--5 of 10 (AIFL protocol) lead times, no difference in RMSE; the benefit of GraphCast is carried by its raw channels\\
GraphCast channels removed, HRES only (test, three seeds) & Worse in \kge{} and NSE at 10/10 lead times in both protocols (operational \kge{}: $-0.015$ at LT1 to $-0.030$ at LT10) and in RMSE at 8/10 (operational) and 10/10 (AIFL protocol) lead times\\
Upper-tail weighted CRPS ($\lambda=1$, $q_{90}$; test, three seeds) & Point-forecast scores unchanged (0/10 significant lead times) and CRPS within 2\%; higher CSI for floods of 10-year return period and above in every seed (Section~\ref{sec:tail}); introduced after the test-period results, hence exploratory\\
\bottomrule
\end{tabular}
\end{table}

\subsection{Probabilistic Skill and Calibration}
\begin{figure}[t]
\centering\includegraphics[width=\textwidth]{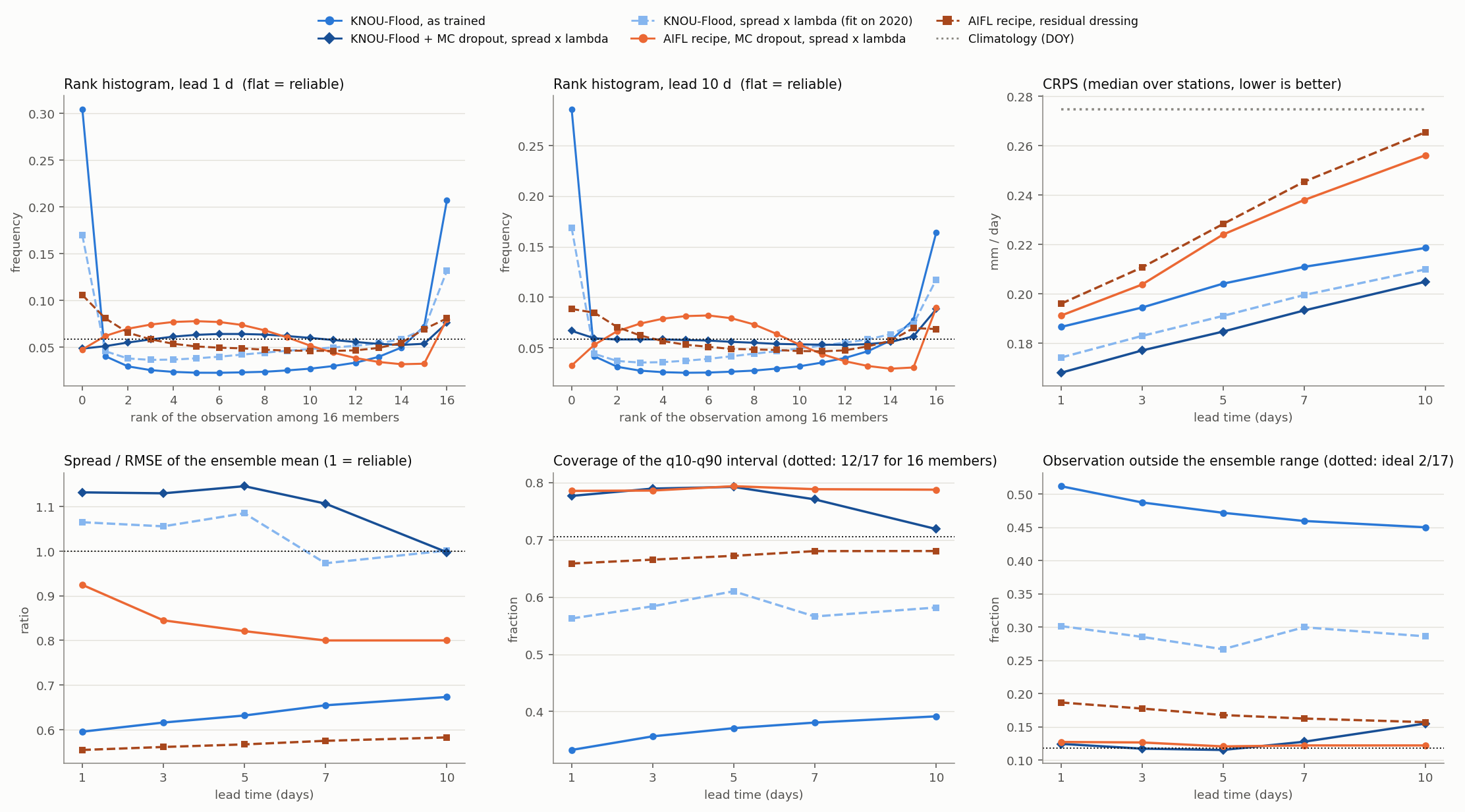}
\caption{Probabilistic forecast quality of \rfc{} and of probabilistic versions of the \recipe{} baseline (test, three seeds). Top: rank histograms (LT1 and LT10) and CRPS. Bottom: spread--error ratio, coverage of the q10--q90 interval (dotted: 12/17 for 16 members) and fraction of observations outside the ensemble range (dotted: 2/17). The spread factor $\lambda_j$ and the dressing residuals are fitted on the validation year 2020 only; point forecasts (ensemble means) are unchanged by the spread scaling.}\label{fig:prob}
\end{figure}

Table~\ref{tab:prob} and Figure~\ref{fig:prob} compare the ensembles. \rfc{} has lower CRPS than both fair probabilistic versions of the baseline at every lead time: the paired intervals of the CRPS difference exclude zero, and the difference is in favor of \rfc{} at 78--93\% of the stations. At LT10 the probabilistic baselines are no better than climatology (CRPSS $-0.01$), whereas \rfc{} reaches $+0.21$. As trained, the ensemble is clearly under-dispersive: the observation falls outside the range of the 16 members in 45--51\% of the cases (ideal 11.8\%). Dropout on its own is far from sufficient: without scaling, the Monte Carlo dropout ensemble of the baseline is strongly under-dispersive (spread/RMSE 0.15--0.23, and 60--65\% of the observations outside the ensemble range). Scaling the deviations alone ($\lambda$) corrects the spread--error ratio but not the tails; adding dropout at inference almost calibrates the ensemble (12--16\% outside, nearly flat rank histogram) and lowers the CRPS further, without retraining and without changing the point forecast. The interval coverage of the final system is slightly above the ideal value of 12/17 at short lead times.

\begin{table}[t]
\centering\small
\caption{Probabilistic scores (test, three seeds, median over stations, 16 members). CRPS in mm\,d$^{-1}$; ``outside'' is the fraction of observations outside the ensemble range (ideal 0.118); the ideal spread/RMSE ratio is 1. $\lambda$ and the dressing residuals are fitted on 2020 only.}\label{tab:prob}
\begin{tabular}{lccccc}
\toprule
CRPS & LT1 & LT3 & LT5 & LT7 & LT10\\
\midrule
\rfc, as trained & 0.187 & 0.194 & 0.204 & 0.211 & 0.219\\
\rfc{} + MC dropout, spread $\times\lambda$ & \textbf{0.168} & \textbf{0.177} & \textbf{0.185} & \textbf{0.193} & \textbf{0.205}\\
\recipe, MC dropout, spread $\times\lambda$ & 0.191 & 0.204 & 0.224 & 0.238 & 0.256\\
\recipe, residual dressing & 0.196 & 0.211 & 0.228 & 0.245 & 0.265\\
Climatology & 0.275 & 0.275 & 0.275 & 0.275 & 0.275\\
\midrule
Outside range (LT1 / LT10): \rfc{} as trained & \multicolumn{5}{l}{0.512 / 0.450}\\
\phantom{Outside range (LT1 / LT10): }\rfc{} + MC dropout, $\times\lambda$ & \multicolumn{5}{l}{0.124 / 0.155}\\
\phantom{Outside range (LT1 / LT10): }\recipe{} MC dropout, $\times\lambda$ & \multicolumn{5}{l}{0.127 / 0.122}\\
Spread/RMSE (LT1 / LT10): \rfc{} as trained; \rfc{} + MC, $\times\lambda$ & \multicolumn{5}{l}{0.60 / 0.67; 1.13 / 1.00}\\
\bottomrule
\end{tabular}
\end{table}

\subsection{Flood Event Performance}
Beyond overall temporal accuracy, operational flood forecasting depends on the reliable detection of high-flow events. Table~\ref{tab:events} reports the event-based verification of the ensemble mean at LT1 under the dual- and single-threshold frameworks (Section~\ref{sec:protocols}); the statistics pool all stations and test days of the three seeds, and \rfc{} is compared with the deterministic baseline. Under the single-threshold framework \rfc{} detects more events than the baseline at all return periods (POD 0.40 against 0.38 at 1.5 years, 0.19 against 0.18 at 5 years and 0.15 against 0.13 at 10 years) and produces fewer false alarms (FAR 0.38 against 0.48 at 1.5 years and 0.81 against 0.84 at 10 years). Under the dual-threshold framework the gain lies in the POD (0.47 against 0.40 at 1.5 years) at a similar FAR (0.53 against 0.54). As in AIFL, the single-threshold framework exposes substantially higher false alarm ratios for rare events, and the POD decreases with event rarity. Our verification pools all station-days with an exact-day criterion, and the numbers are not directly comparable with those reported for AIFL, whose event counts for rare return periods are very small.

\begin{table}[t]
\centering\small
\caption{Global event-based verification statistics at LT1 under the dual-threshold and single-threshold frameworks (test; three seeds pooled). Dual threshold: forecast exceedances use Gumbel thresholds fitted to the annual maxima of the model's own ERA5-Land-driven simulation, observed exceedances those of the observations. Single threshold: both use thresholds fitted to the observed annual maxima.}\label{tab:events}
\begin{tabular}{lcccccc}
\toprule
 & \multicolumn{2}{c}{FAR} & \multicolumn{2}{c}{POD} & \multicolumn{2}{c}{Precision}\\
Return period & Dual & Single & Dual & Single & Dual & Single\\
\midrule
\multicolumn{7}{l}{\emph{\rfc{} (ensemble mean)}}\\
1.5 yr & 0.534 & 0.381 & 0.470 & 0.398 & 0.466 & 0.619 \\
2 yr & 0.583 & 0.423 & 0.358 & 0.315 & 0.417 & 0.577 \\
5 yr & 0.695 & 0.702 & 0.251 & 0.186 & 0.305 & 0.298 \\
10 yr & 0.685 & 0.811 & 0.274 & 0.150 & 0.315 & 0.189 \\
20 yr & 0.692 & 0.885 & 0.298 & 0.119 & 0.308 & 0.115 \\
50 yr & 0.781 & 0.962 & 0.283 & 0.064 & 0.219 & 0.038 \\
\midrule
\multicolumn{7}{l}{\emph{\recipe{} baseline (deterministic)}}\\
2 yr & 0.587 & 0.513 & 0.306 & 0.297 & 0.413 & 0.487 \\
5 yr & 0.693 & 0.730 & 0.236 & 0.184 & 0.307 & 0.270 \\
10 yr & 0.732 & 0.844 & 0.231 & 0.133 & 0.268 & 0.156 \\
20 yr & 0.748 & 0.901 & 0.235 & 0.108 & 0.252 & 0.099 \\
50 yr & 0.800 & 0.974 & 0.225 & 0.045 & 0.200 & 0.026 \\
\bottomrule
\end{tabular}
\end{table}

\begin{figure}[t]
\centering\includegraphics[width=\textwidth]{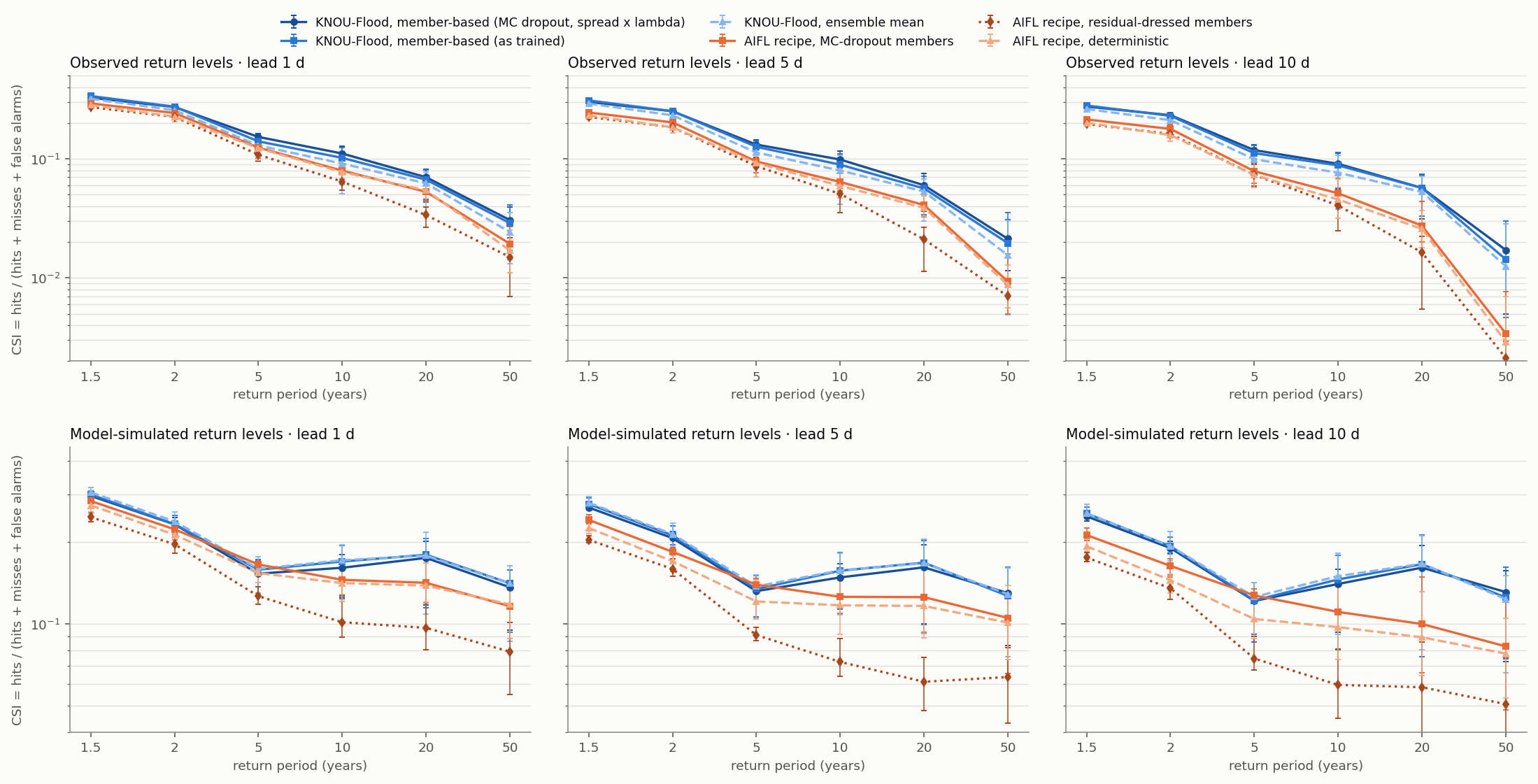}
\caption{Critical success index versus return period for deterministic, ensemble-mean and member-based event forecasts at LT1, LT5 and LT10 under the single-threshold (top) and dual-threshold (bottom) frameworks, pooled over stations and days of the test period with exact-day matching. A member-based event is forecast when at least a fraction $p^{*}$ of the 16 members exceeds the return level; $p^{*}$ and $\lambda_j$ are tuned on 2020 only. Bars show the minimum and maximum over the three seeds.}\label{fig:events}
\end{figure}

Because \rfc{} produces an ensemble, an event can also be forecast from the fraction of members exceeding the threshold. Table~\ref{tab:members} and Figure~\ref{fig:events} show the CSI of member-based event forecasts against ensemble-mean and deterministic forecasts. For return periods of 1.5 and 2 years (observed return levels), the member-based forecasts of \rfc{} have a higher CSI than the ensemble mean and than every baseline system (deterministic, MC dropout, residual dressing) at every lead time and in every seed; the exceedance fraction of the members is therefore informative beyond the mean. For return periods of five years and above the picture depends on the seed: in two seeds the CSI of \rfc{} is about twice that of the best baseline (for example LT5 and 10 years: 0.116 and 0.111 against 0.054 and 0.057), in the third it is 25--35\% lower (0.065 against 0.087). Pooled over the seeds \rfc{} is ahead in every cell of Table~\ref{tab:members}, but we do not claim superiority for extreme events. Under the dual-threshold framework the results for 5--50 years are inconclusive.

\begin{table}[t]
\centering\small
\caption{Critical success index for observed return levels at LT5, pooled over three seeds ($p^*$ tuned on 2020).}\label{tab:members}
\begin{tabular}{lcccccc}
\toprule
Return period (years) & 1.5 & 2 & 5 & 10 & 20 & 50\\
\midrule
\rfc, ensemble mean & 0.291 & 0.232 & 0.113 & 0.080 & 0.053 & 0.016\\
\rfc{} + MC dropout, member-based & \textbf{0.300} & \textbf{0.251} & \textbf{0.132} & \textbf{0.099} & \textbf{0.060} & \textbf{0.021}\\
\recipe, deterministic & 0.232 & 0.184 & 0.092 & 0.059 & 0.039 & 0.009\\
\recipe, MC dropout, member-based & 0.245 & 0.202 & 0.095 & 0.064 & 0.041 & 0.009\\
\bottomrule
\end{tabular}
\end{table}

\paragraph{Tail-weighted training.}\label{sec:tail} The seed dependence for rare floods led us to add an upper-tail term to the fine-tuning loss: the CRPS of the ensemble after the transformation $x\mapsto\max(x,t_b)$, a threshold-weighted CRPS \citep{gneiting2011}, where $t_b$ is the 90th percentile of the 1980--2015 discharge of each station, added to the fair CRPS with weight one. Both settings were fixed in advance and not tuned, but the idea was prompted by the test-period result above, so we treat it as exploratory. The variant leaves the point-forecast scores unchanged (no significant difference in \kge{}, NSE or RMSE at any lead time in either protocol) and the CRPS within 2\%. For floods of 10-year return period and above, its CSI is higher than that of the standard model in every cell and every seed (45 of 45 combinations of lead time, return period and seed; Table~\ref{tab:tail}). Relative to the best baseline system it is ahead in all cells in two seeds and in 9 of 15 cells in the third, where it remains slightly behind at the shortest lead times for 10- and 50-year events. At 5 years the improvement over the standard model is small, and in the third seed the baseline remains ahead.

\begin{table}[t]
\centering\footnotesize
\caption{CSI of member-based event forecasts (MC dropout, spread $\times\lambda$, $p^*$ tuned on 2020) for observed return levels, mean over three seeds: standard \rfc, \rfc{} trained with the tail-weighted CRPS (bold) and the best baseline system of each seed.}\label{tab:tail}
\begin{tabular}{lccccccccc}
\toprule
 & \multicolumn{3}{c}{LT1} & \multicolumn{3}{c}{LT5} & \multicolumn{3}{c}{LT10}\\
Return period & Std. & Tail & Base & Std. & Tail & Base & Std. & Tail & Base\\
\midrule
1.5 yr & 0.326 & \textbf{0.332} & 0.292 & 0.300 & \textbf{0.306} & 0.245 & 0.275 & \textbf{0.282} & 0.214 \\
2 yr & 0.272 & \textbf{0.280} & 0.243 & 0.251 & \textbf{0.260} & 0.204 & 0.232 & \textbf{0.245} & 0.179 \\
5 yr & 0.153 & \textbf{0.157} & 0.125 & 0.131 & \textbf{0.139} & 0.100 & 0.117 & \textbf{0.130} & 0.086 \\
10 yr & 0.111 & \textbf{0.127} & 0.081 & 0.097 & \textbf{0.116} & 0.066 & 0.089 & \textbf{0.115} & 0.053 \\
20 yr & 0.070 & \textbf{0.092} & 0.055 & 0.059 & \textbf{0.081} & 0.041 & 0.054 & \textbf{0.083} & 0.030 \\
50 yr & 0.030 & \textbf{0.047} & 0.022 & 0.021 & \textbf{0.035} & 0.011 & 0.015 & \textbf{0.035} & 0.006 \\
\bottomrule
\end{tabular}
\end{table}

\section{Conclusion and Future Directions}\label{sec:conclusion}
This study presents \rfc, a machine learning model for global streamflow forecasting that extends the AIFL recipe with a longer input window, lead-aware inputs from two forecast sources and a probabilistic output. Under identical data, stations, splits and metrics, \rfc{} improves the median \kge{}, NSE and RMSE of an AIFL-recipe baseline at all ten lead times, by 0.04 in \kge{} at LT1 and 0.14 at LT10 under the operational protocol, and the improvement survives a larger baseline and basins unseen in training. We show that a large part of the lead-time decay of the published protocol is an artifact of how forecasts are fed into the model, so that skill at long lead times is higher than the protocol suggests. The ensemble reaches a lower CRPS than fair probabilistic versions of the baseline and, with dropout at inference, is nearly calibrated without retraining. Event-based verification indicates reliable detection of floods up to the two-year return period, whereas the benefit of the standard model for rarer extremes depends on the training seed. A variant trained with a tail-weighted CRPS raises the detection of floods of 10-year return period and above in every seed without changing point or probabilistic skill, although it was introduced after the test-period results and remains exploratory. This performance profile reflects a model that improves the bulk of the distribution and the probabilistic information, rather than the extremes.

Several avenues remain for future development. The AIFL-recipe baseline is a re-implementation trained with about 18\% of the published number of updates and scores below the published values; training both models at the published budget would clarify how much of the gain persists. The extreme-event results need more seeds and an independent test period, in particular for the tail-weighted objective, which may reduce the underestimation of peaks. The model treats stations independently; coupling stations along the river network would allow the propagation of flow to be represented. Finally, the integration of probabilistic ensemble forcing, of near-real-time discharge observations and of multi-source precipitation products could refine event detection and support risk-based decision-making. By providing a lead-aware and probabilistic architecture that balances simplicity with competitive skill, \rfc{} establishes a reproducible baseline for global streamflow forecasting.

\appendix
\section{Case Studies}
\begin{figure}[!htbp]
\centering\includegraphics[width=0.85\textwidth]{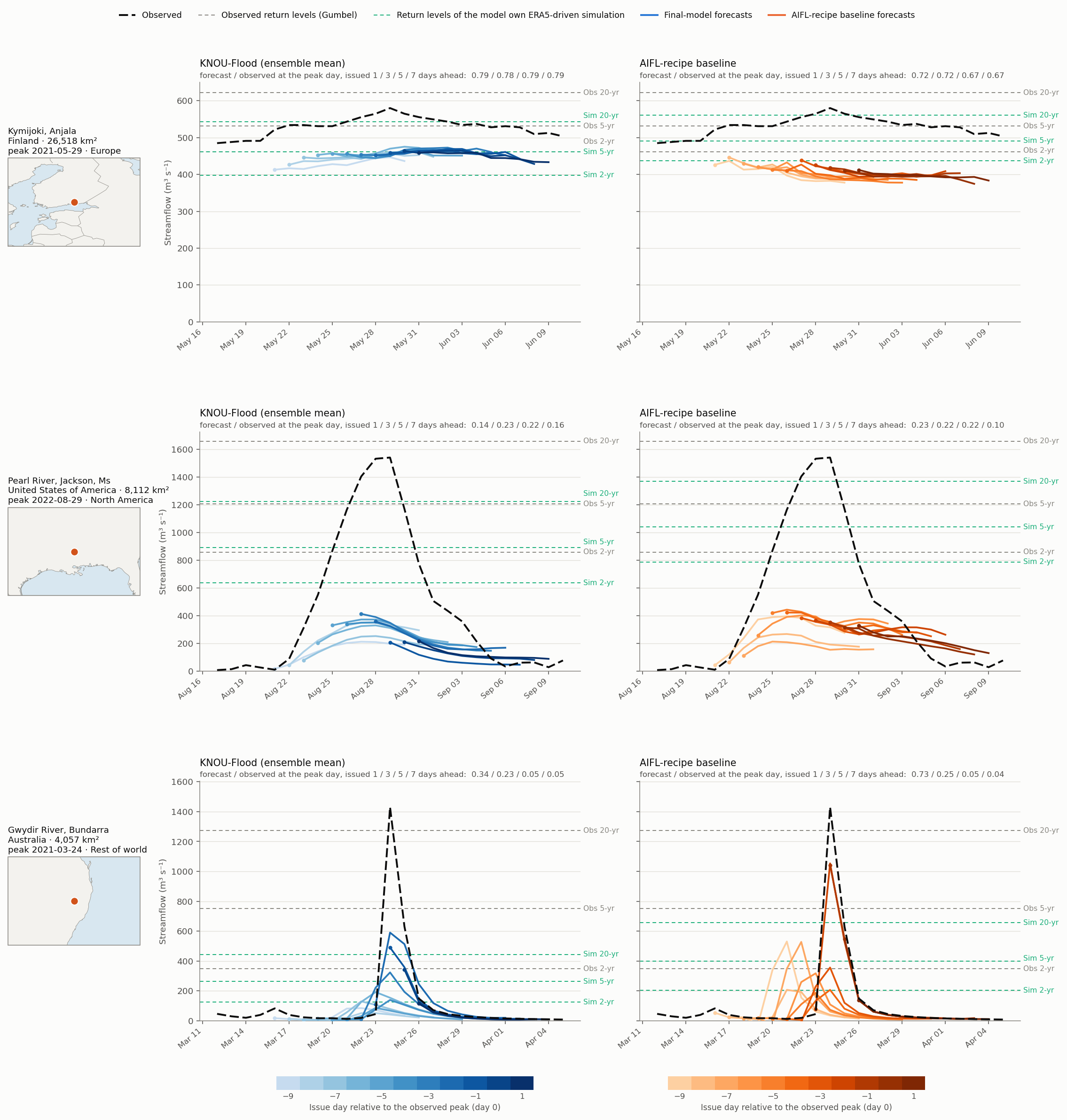}
\caption{Randomly drawn flood events (one per region, observed peak above the 10-year level): all 10-day forecasts issued from 9 days before to 1 day after the observed peak, colored by issue day, with observed and model-simulated return levels. Events are drawn at random from all observed test-period peaks above the observed 10-year level, not selected on model skill; forecasts are ensemble means of one seed under the operational protocol.}\label{fig:cases}
\end{figure}

\begin{figure}[!htbp]
\centering\includegraphics[width=0.95\textwidth]{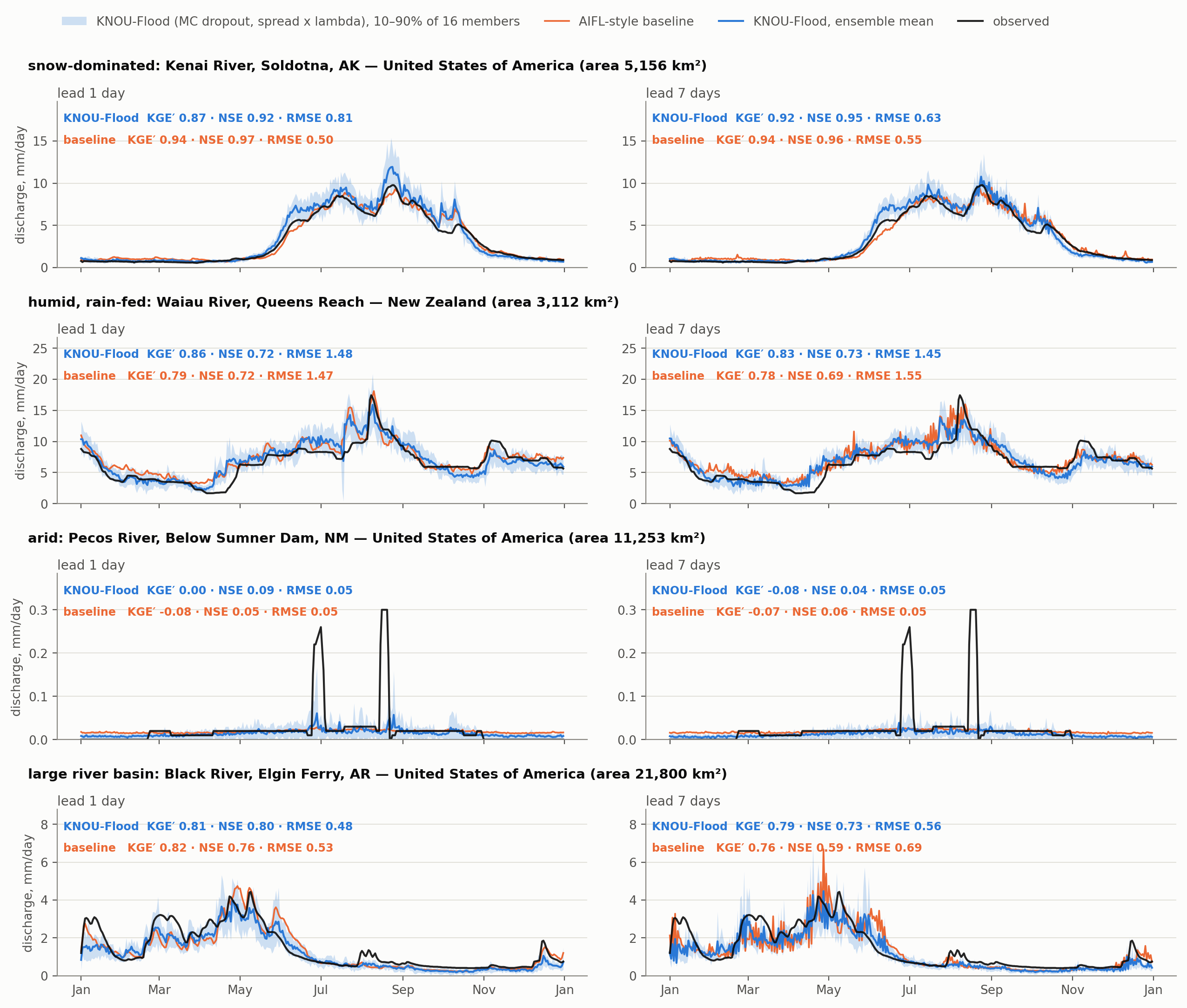}
\caption{Randomly drawn stations (one per regime) in 2022 at lead times 1 and 7 days; shading shows the 10--90\% range of the 16 members. Scores refer to the plotted year. For the arid station flows stay below 0.3 mm\,d$^{-1}$, so its scores are not meaningful.}\label{fig:gauges}
\end{figure}

\begin{figure}[!htbp]
\centering\includegraphics[width=0.9\textwidth]{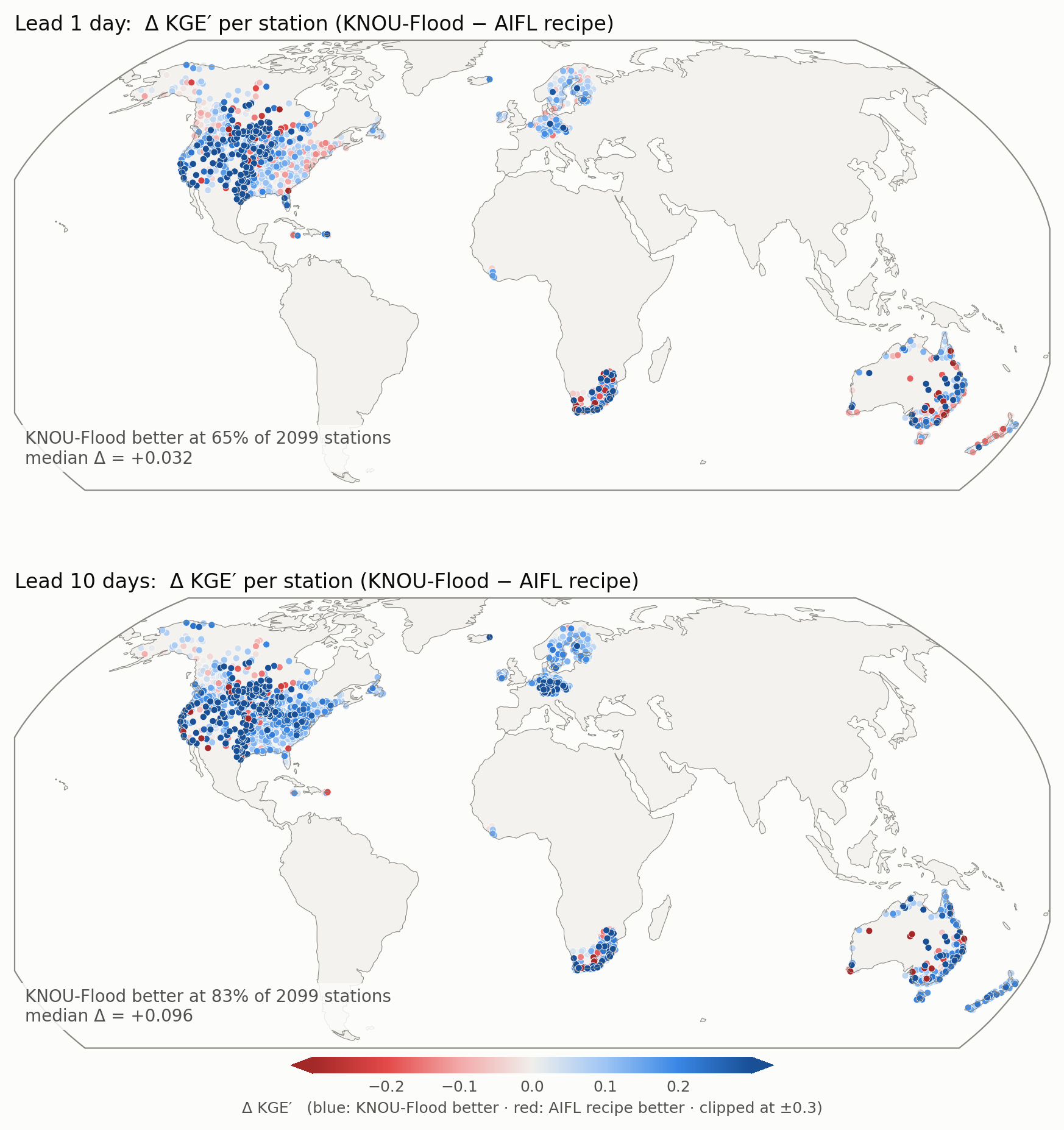}
\caption{Per-station difference in \kge{} between \rfc{} and the \recipe{} baseline at LT1 and LT10 (operational protocol, mean of three seeds).}\label{fig:dmap}
\end{figure}

While aggregate statistics summarize global test-set performance, they do not fully convey operational realism during individual events. Figure~\ref{fig:cases} shows three flood events drawn at random (one per region) from all test-period observed peaks above the observed 10-year level; the events were not selected on model skill. Both models underestimate the peaks, and in one case the baseline is clearly closer at LT1. Figure~\ref{fig:gauges} shows one randomly drawn station per hydrological regime in 2022, and Figure~\ref{fig:dmap} the difference in \kge{} between \rfc{} and the baseline at each station.

\clearpage

\section*{Code and Data Availability}
Caravan, GRDC-Caravan and Caravan-MultiMet are available via Zenodo and Google Cloud Storage as described in the original publications. The code, trained weights, per-station results and the commands used to produce the results of this paper are available at \url{https://github.com/jmj2316/KNOU-Flood}.

\section*{Author Contributions}
Following the CRediT taxonomy: Minjong Cheon: Conceptualization, Methodology, Software, Validation, Formal analysis, Investigation, Data curation, Visualization, Writing -- original draft, Writing -- review \& editing.

\end{document}